\PassOptionsToPackage{unicode}{hyperref}
\PassOptionsToPackage{hyphens}{url}
\PassOptionsToPackage{dvipsnames,svgnames,x11names}{xcolor}
\documentclass[
  letterpaper,
  DIV=11,
  numbers=noendperiod]{scrartcl}

\usepackage{amsmath,amssymb}
\usepackage{iftex}
\ifPDFTeX
  \usepackage[T1]{fontenc}
  \usepackage[utf8]{inputenc}
  \usepackage{textcomp} 
\else 
  \usepackage{unicode-math}
  \defaultfontfeatures{Scale=MatchLowercase}
  \defaultfontfeatures[\rmfamily]{Ligatures=TeX,Scale=1}
\fi
\usepackage{lmodern}
\ifPDFTeX\else  
\fi
\IfFileExists{upquote.sty}{\usepackage{upquote}}{}
\IfFileExists{microtype.sty}{
  \usepackage[]{microtype}
  \UseMicrotypeSet[protrusion]{basicmath} 
}{}
\makeatletter
\@ifundefined{KOMAClassName}{
  \IfFileExists{parskip.sty}{%
    \usepackage{parskip}
  }{
    \setlength{\parindent}{0pt}
    \setlength{\parskip}{6pt plus 2pt minus 1pt}}
}{
  \KOMAoptions{parskip=half}}
\makeatother
\usepackage{xcolor}
\ifx\paragraph\undefined\else
  \let\oldparagraph\paragraph
  \renewcommand{\paragraph}[1]{\oldparagraph{#1}\mbox{}}
\fi
\ifx\subparagraph\undefined\else
  \let\oldsubparagraph\subparagraph
  \renewcommand{\subparagraph}[1]{\oldsubparagraph{#1}\mbox{}}
\fi

\providecommand{\tightlist}{%
  \setlength{\itemsep}{0pt}\setlength{\parskip}{0pt}}\usepackage{longtable,booktabs,array}
\usepackage{calc} 
\usepackage{etoolbox}
\makeatletter
\patchcmd\longtable{\par}{\if@noskipsec\mbox{}\fi\par}{}{}
\makeatother
\IfFileExists{footnotehyper.sty}{\usepackage{footnotehyper}}{\usepackage{footnote}}
\makesavenoteenv{longtable}
\usepackage{graphicx}
\makeatletter
\def\maxwidth{\ifdim\Gin@nat@width>\linewidth\linewidth\else\Gin@nat@width\fi}
\def\maxheight{\ifdim\Gin@nat@height>\textheight\textheight\else\Gin@nat@height\fi}
\makeatother
\setkeys{Gin}{width=\maxwidth,height=\maxheight,keepaspectratio}
\makeatletter
\def\fps@figure{htbp}
\makeatother
\newlength{\cslhangindent}
\newlength{\csllabelwidth}
\newlength{\cslentryspacingunit} 
\newenvironment{CSLReferences}[2] 
 {
  \setlength{\parindent}{0pt}
  \ifodd #1
  \let\oldpar\par
  \def\par{\hangindent=\cslhangindent\oldpar}
  \fi
  \setlength{\parskip}{#2\cslentryspacingunit}
 }%
 {}
\usepackage{calc}

\KOMAoption{captions}{tableheading}
\makeatletter
\@ifpackageloaded{caption}{}{\usepackage{caption}}
\AtBeginDocument{%
\ifdefined\contentsname
  \renewcommand*\contentsname{Table of contents}
\else
  \newcommand\contentsname{Table of contents}
\fi
\ifdefined\listfigurename
  \renewcommand*\listfigurename{List of Figures}
\else
  \newcommand\listfigurename{List of Figures}
\fi
\ifdefined\listtablename
  \renewcommand*\listtablename{List of Tables}
\else
  \newcommand\listtablename{List of Tables}
\fi
\ifdefined\figurename
  \renewcommand*\figurename{Figure}
\else
  \newcommand\figurename{Figure}
\fi
\ifdefined\tablename
  \renewcommand*\tablename{Table}
\else
  \newcommand\tablename{Table}
\fi
}
\@ifpackageloaded{float}{}{\usepackage{float}}
\floatstyle{ruled}
\@ifundefined{c@chapter}{\newfloat{codelisting}{h}{lop}}{\newfloat{codelisting}{h}{lop}[chapter]}
\floatname{codelisting}{Listing}

\makeatother
\makeatletter
\@ifpackageloaded{caption}{}{\usepackage{caption}}
\@ifpackageloaded{subcaption}{}{\usepackage{subcaption}}
\makeatother
\makeatletter
\@ifpackageloaded{tcolorbox}{}{\usepackage[skins,breakable]{tcolorbox}}
\makeatother
\makeatletter
\@ifundefined{shadecolor}{\definecolor{shadecolor}{rgb}{.97, .97, .97}}
\makeatother
\ifLuaTeX
  \usepackage{selnolig}  
\fi
\IfFileExists{bookmark.sty}{\usepackage{bookmark}}{\usepackage{hyperref}}
\IfFileExists{xurl.sty}{\usepackage{xurl}}{} 
\hypersetup{
  pdftitle={Generative models for visualising abstract social processes},
  pdfauthor={Aleksi Knuutila},
  colorlinks=true,
  linkcolor={blue},
  filecolor={Maroon},
  citecolor={Blue},
  urlcolor={Blue},
  pdfcreator={LaTeX via pandoc}}

\title{Generative models for visualising abstract social processes}
\usepackage{etoolbox}
\makeatletter
\providecommand{\subtitle}[1]{
  \apptocmd{\@title}{\par {\large #1 \par}}{}{}
}
\makeatother
\subtitle{Guiding streetview image synthesis of StyleGAN2 with indices
of deprivation}
\author{Aleksi Knuutila}
\date{2023-09-25}

\begin{document}
\maketitle
\begin{abstract}
This paper presents a novel application of Generative Adverserial
Networks (GANs) to study visual aspects of social processes. I train a a
StyleGAN2-model on a custom dataset of 14,564 images of London, sourced
from Google Streetview taken in London. After training, I invert the
images in the training set, finding points in the model's latent space
that correspond to them, and compare results from three inversion
techniques. I connect each data point with metadata from the Indices of
Multiple Deprivation, describing income, health and environmental
quality in the area where the photographs were taken. It is then
possible to map which parts of the model's latent space encode visual
features that are distinctive for health, income and environmental
quality, and condition the synthesis of new images based on these
factors. The synthetic images created reflect visual features of social
processes that were previously unknown and difficult to study,
describing recurring visual differences between deprived and privileged
areas in London. GANs are known for their capability to produce a
continuous range of images that exhibit visual differences. The paper
tests how to exploit this ability through visual comparisons in still
images as well as through an interactive website where users can guide
image synthesis with sliders. Though conditioned synthesis has its
limitations and the results are difficult to validate, the paper points
to the potential for generative models to be repurposed to be parts of
social scientific methods.
\end{abstract}
\ifdefined\Shaded\renewenvironment{Shaded}{\begin{tcolorbox}[breakable, boxrule=0pt, interior hidden, sharp corners, borderline west={3pt}{0pt}{shadecolor}, frame hidden, enhanced]}{\end{tcolorbox}}\fi

\hypertarget{introduction}{%
\subsection{Introduction}\label{introduction}}

Social scientists and urban planners have long studied how urban forms
and landscapes in cities reflect but also perpetuate social distinctions
and inequalities. Different areas of cities vary in demographic factors,
such as their residents' income levels and ethnic background.
Additionally, they may diverge in terms of perceived safety,
cleanliness, historical significance, and vibrancy, among many other
potential dimensions of evaluation (Nasar 1998). Since cities produce
quantitative data of numerous sorts that can be spatially joined,
city-internal inequality is frequently described through statistical
comparison and correlation. Yet recent scholarship in the inequality
domain has also sought to foreground aspects of inequality that are not
so clearly captured by numerical aggregation (Dorling 2012), such as
disparities in experiences or perceptions of city spaces (Salesses,
Schechtner, and Hidalgo 2013) or how particular spaces in the city
symbolise and maintain differences in social capital (Tonkiss 2015).
This paper describes an application of image synthesis to better
understand how the visual appearances of cities and inequality connect.

In the past decade, a significant amount of scholarship has explored how
large-scale datasets of images from cities could be utilised in
research. In particular, the Google Street View service has made
millions of street-level photographs publicly available. Many studies
using street-view images have examined the predictive potential of
visual data from cities and have modelled how these images may predict
non-visual attributes of cities (Biljecki and Ito 2021). For example,
street-level photographs prove sufficient for models to estimate levels
of crime (Arietta et al. 2014), health (Kang et al. 2020), and property
valuation (Johnson, Tidwell, and Villupuram 2020). Some researchers also
attempt to interpret the functioning of the predictive models and assess
which visual features affect their predictions. Quercia et al., for
instance, identified sets of ``visual words'' -- salient areas in the
input images that have the strongest effect on the model's
classification (2014). In some cases, however, predictive models might
operate with features that are not easy for humans to understand (Kim et
al. 2018, 1), and the results from saliency-type analysis does not
provide a fine-grained visual explanation of the phenomena under study
(Goetschalckx et al. 2019, 1).

This paper presents a novel framework for modelling the visual aspects
of urban processes. A generative model was trained on street-level
photographs to create synthetic images of cities. The generation of
these images is conditioned by socioeconomic data related to income,
health, and education. For the demonstration in the paper a StyleGAN2
model, a popular architecture for generative adversarial networks
(GANs), was applied. It was trained on a custom dataset of 14,564 images
captured in London from Google Street View. The technique exploits GANs'
ability to generate a continuum of images with fine-granularity
differences in their visual attributes.

The project took advantage of the capacity of GANs to edit selected
visual features independently of other features. Typically, in
applications of GANs, the visual features that need to be changed in the
images are known \emph{a priori}. GANs function as a practical means of
operationalising such changes. For example, StyleGAN2 models are often
applied for image manipulation in domains such as human faces. For the
demonstration provided here, the project applied popular image
manipulation techniques but targeted abstract, social variables for
which the visual correlates are unknown. Hence, the visualisations
produced by GANs produce novel insights into the effects of deprivation
and privilege on the street-level appearance of the city, resulting in
what Goetschalckx et al.~have called the ``visual definition'' of these
phenomena (2019).

The paper presents the results of this visualisation based on still
images. Furthermore, the project includes a a web-based interface
through which users can guide the image synthesis by selecting input
values in three dimensions (income, health, and education) on a sliding
scale. This interface highlights the potential for interactivity in
visualisation, especially when the method involves models such as GANs,
which produce a continuous so-called latent space corresponding to
meaningful generated images.

\hypertarget{related-work}{%
\subsection{Related work}\label{related-work}}

\hypertarget{image-editing-with-gans}{%
\subsubsection{Image editing with GANs}\label{image-editing-with-gans}}

Many architectures and types of models exist for image generation, among
which contrastive learning and diffusion models have yielded
particularly good results in recent years, especially in the
text-to-image domain (Elasri et al. 2022). The GAN architecture has
existed for some time but is still regarded as in the least close to
state-of-the-art for high-resolution image synthesis with medium-sized
training sets (Gui et al. 2021). Their approach is based on identifying
the lower-dimensionality manifold in which images from the training set
are concentrated (Lei et al. 2020). The model maps data points drawn
from a Gaussian distribution from a latent space onto the patterns
observed in the images. The central benefit of this model type is that
it is possible to interpolate between two points in the latent space and
obtain continuous changes in the images generated (Shen et al. 2020,
9245).

A critical property of the latent space is its ability to capture
semantic attributes of images and organise them into more or less
disentangled subspaces without supervision. This means that distinct
characteristics or features of the generated data can be controlled or
modified independently by moving along specific directions in the latent
space. For example, algorithms for editing images by traversing the
latent space have been proposed that alter specific image attributes
while holding others constant by moving along carefully chosen
latent-space directions. The traversal methods typically assume scalar
variables of interest that can be modelled with linear functions -- for
instance, linear regressions or support-vector machines (Liu et al.
2023). This approach is appropriate for attributes such as gender,
facial expression, age, or hair colour when editing faces or, more
generally, continuous visual transformations such as colour changes,
camera movement, or change in position. For instance, InterfaceGAN,
underpinning the approach described in this paper, fits a support-vector
machine to find a separation boundary within the latent space for a
dichotomous variable and uses the orthogonal subspaces as directions for
traversal (Shen et al. 2020).

GANs typically work by mapping random vectors to generated images. For
image editing to be possible with arbitrarily chosen images, it is
necessary to identify points in the model's latent space corresponding
to the selected input images through an inversion process. In their
review of inversion methods, Xia et al.~distinguish between
optimisation- and learning-based methods (2022). Optimisation-based
methods formulate the inversion problem as an optimisation task. They
define an objective function (loss function) that quantifies the
difference between the generated and the target image, then iteratively
adjusts the latent vector to minimise that loss function's value.
Learning-based methods involve training new models, often called
``inverter'' networks or encoder-decoder networks. These models learn to
map images from the data distribution back to the latent space. Xia et
al.~posited that a trade-off exists between visual fidelity and
editability of GAN inversion, and advocates of some learning-based
methods claim that these improve the editability of images while
maintaining high fidelity.

\hypertarget{gans-for-the-study-of-unknown-visual-features}{%
\subsubsection{GANs for the study of unknown visual
features}\label{gans-for-the-study-of-unknown-visual-features}}

Many GAN-based image-editing applications focus on manipulating images
by proceeding from visual features that are known \emph{a priori}. The
most common realm for image synthesis, though not the only one, is the
human face, which is manipulated based on gender, age, etc. However,
some applications of GANs are employed to visualise at least partially
unknown visual attributes through the generation of images. In these
cases, the synthesis of new images is handled such that it is controlled
and conditioned by some outside variables. The resulting images
themselves contain some new information about the domain under study.
While some research has used GANs to describe unknown visual properties,
a literature review has not revealed any that has focused on deprivation
in cities or conditioning image synthesis by means of the technique
presented here.

One example of the general approach is the work by Schmidt et al., who
trained a CycleGAN model on street-view images of houses before and
after extreme weather events (2019). The model thus learns a mapping
that could be applied to images of locations that have not yet
experienced such events. Their project aimed to form an intuitive
understanding of the effects of climate change, enabling individuals to
make more informed choices about their climate future. Another example
from the urban domain is a study in which Langer et al.~focused on the
impact of climate change on population distribution and land use, as
revealed by satellite images (2020). The researchers developed a
generative model framework called SCALAE, a spatially conditional
version of the pre-existing ALAE architecture. This framework afforded
to generate satellite imagery based on gridded population distributions.
Through explicit disentangling of the population from the model's latent
space, custom population forecasts could be fed in for the generated
imagery. This approach facilitated estimating land cover and land-use
changes and provided for realistic visualisation of expected local
changes due to climate change. Modelling causal processes with GANs has
also received attention in the natural sciences (Chen et al. 2022). For
example, Osokin et al.~overcame limits in cellular imaging by developing
conditional GANs that model the process of protein localisation in yeast
cells (2017).

With their GANalyze model, Goetschalckx and colleagues explored the use
of GANs for understanding and setting cognitive properties of images.
The authors began by training a GAN on a large image dataset to achieve
this. Then, the images generated were evaluated by human participants,
who performed recall-centred tasks with the images in varying experiment
conditions. The feedback from human participants can direct the
synthesis of images, thereby leading toward a ``visual definition of
image memorability''.

\hypertarget{visual-comparisons-as-forms-of-visualisation}{%
\subsubsection{Visual comparisons as forms of
visualisation}\label{visual-comparisons-as-forms-of-visualisation}}

This paper experimented with displaying generated images that facilitate
comparisons through still images and interactive interfaces. Among other
scholars, Edward Tufte has described the design of such comparisons. In
his \emph{Visual Explanations}, Tufte discussed the multiple ways
visualisations can facilitate comparisons in what he calls visual
parallelism (1998, 80--83). Parallelism involves repetition and
contrast, the most basic form of which might be ``before/after'' images,
displaying divergent views of the same subject. Parallelism between
images can be achieved in several quite different ways. When images are
close to each other, they form parallelism in space, in contrast against
parallelism in time, contrasting a remembered image with the image
currently being viewed. According to Tufte, appropriately chosen forms
of parallelism can ``reveal repetition and change, pattern and surprise
-- the defining elements in the idea of information'' (1998, p.~105).

Another taxonomy of comparative designs presented by Gleicher et
al.~includes three main categories: juxtaposition, superposition, and
explicit encoding of relationships. Juxtaposition involves showing
different objects separately, often in small multiples or side-by-side
views. Conversely, superposition involves overlaying objects in a single
space, either by making one of them semi-transparent or by using
explicit encodings to emphasise patterns in the overlaid views. Finally,
direct encoding of relationships involves abstracting complex objects
into a superposition view and explicitly encoding the relations between
them.

\hypertarget{data}{%
\subsection{Data}\label{data}}

\hypertarget{the-street-view-dataset}{%
\subsubsection{The Street View dataset}\label{the-street-view-dataset}}

The training set of Google Street View images was downloaded via
Google's API in July 2018. Images were sought within a radius of 15
kilometres from Trafalgar Square (all within the administrative area of
the Greater London Authority). Only images representing the view
directly to the left of the direction the vehicle was facing were
selected (i.e., the images captured the buildings on the same side of
the street as the car). The resulting dataset consisted of 538,148
images.

Since these street-view images exhibited a considerable variety of
visual content, a sample from the full dataset was selected,
sufficiently uniform to be represented through a GAN. Among the
selection criteria were the following:

\begin{itemize}
\tightlist
\item
  The image had to contain buildings (as opposed to, for instance, an
  open field).
\item
  The image could not feature a street.
\item
  Images of high-rise buildings had to be removed since these occurred
  at lower frequencies and displayed great variation.
\item
  There were to be no large obstructions of the buildings in view, such
  as scaffolding.
\item
  The photograph had to be oriented largely in alignment with the
  pavement (as opposed to having been taken at an angle).
\end{itemize}

This smaller dataset was produced by manually classifying 1,000 images
and then training a simple visual classifier based on logistic
regression on VGG-16 features to repeat the classification over the full
dataset. The classification results were hand-validated, and images that
did not fit the criteria were removed until the validated dataset
contained 14,564 images. This dataset size has been shown to suffice in
many domains for making a StyleGAN2-ADA, or SG2-ADA, model converge and
produce good results.

\hypertarget{indices-of-multiple-deprivation}{%
\subsubsection{Indices of multiple
deprivation}\label{indices-of-multiple-deprivation}}

Numerical data was also obtained for the relevant area's social and
economic quality for every street-view image. The most geographically
fine-grained data of this type came from the Index of Multiple
Deprivation (IMD), a set of statistical indices used in the United
Kingdom and some other countries to assess and rank areas by level of
socioeconomic deprivation. Based on British administrative data, the
data are aggregated by ``lower-layer super output area''
\texttt{(LSOA)}, where the average population of an LSOA in London in
2010 was 1,722 (Datastore 2022). While the IMD data encompasses multiple
different socioeconomic dimensions, this analysis focuses on three
specific dimensions. These dimensions were selected because they
represent significant aspects of deprivation and privilege. Also, they
are relatively, though not entirely, uncorrelated with each other. These
dimensions and their definition are

\begin{itemize}
\tightlist
\item
  Income deprivation, a measurement that considers the proportion of the
  population in an area with low income, often assessed through several
  indicators (e.g., welfare benefits and tax data)
\item
  Education deprivation, which captures the educational attainment of
  the population, including factors such as the level of academic
  qualifications in the area
\item
  Health deprivation encompasses a range of data reflecting the
  well-being and health status of populations in different regions. Key
  factors include mortality rates, life expectancy, and the prevalence
  of specific health conditions such as mental health issues, chronic
  diseases, and disability.
\end{itemize}

The IMD data used were from 2019 (Statistics 2019). The data are
represented in ordinal rankings, describing the rankings of areas
relative to one another.

\hypertarget{methods}{%
\subsection{Methods}\label{methods}}

\hypertarget{projection}{%
\subsubsection{Projection}\label{projection}}

The aim of the visualisation was to establish a correspondence between
directions in the latent space and the features of the IMD dataset. For
this purpose, it is necessary to apply a process often referred to as
projection, which finds vectors in the latent space that produce images
corresponding as closely as possible to the input ones. As the
description above indicates, there are several methods for performing
projection. For this paper results from three distinct projection
methods were compared: ReStyle (Alaluf, Patashnik, and Cohen-Or 2021),
Encoder4Editing (E4E) (Tov et al. 2021), and the optimisation-based
projector found in the StyleGAN2 repository (Karras et al. 2020).

\hypertarget{finding-semantic-vectors-and-image-editing}{%
\subsubsection{Finding semantic vectors and image
editing}\label{finding-semantic-vectors-and-image-editing}}

Once the projections are done, every street-view image is associated
with both a latent code and additional data characterising deprivation
in the area. The next step is understanding how the visual features
associated with a particular variable are encoded in the model's latent
space. For this purpose, this paper followed the method described by
Shen and colleagues (2020). The socioeconomic IMD data is represented as
a dichotomous class and a hyperplane is identified within the latent
space that functions as a separation boundary delineating the two
classes. The binary classes are formed by selecting the street-view
images in the top and bottom 20\% for each dimension. Lastly, a
support-vector machine is fitted to find the separation boundary between
the two classes. The semantic vectors used to modify the images are the
two directions that are perpendicular to the separation boundary.

The method described by Shen et al.~(2020) makes it possible to edit
images along dimensions simultaneously (e.g.~health and income
simultaneously). To achieve more precise control when manipulating
images, the semantic vectors of different dimensions are made to be
independent from each other with a technique called subspace projection.
Initially, the vectors may be entangled; for instance, areas exhibiting
higher education levels might have visual features associated with
better health. Disentangling the semantic vectors isolates the visual
features specific to educational attainment. For this purpose, we
calculate the projection between two semantic vectors and subtract this
from the original vector. The resulting new semantic vectors are
orthogonal to each other.

\[ \mathbf{n}_1^{o} = \mathbf{n}_1-\left(\mathbf{n}_1^T \mathbf{n}_2\right) \mathbf{n}_2 \]

Once the semantic vectors have been determined, the images are edited by
moving in the vector's direction from the given position or away from it
in the latent space. Latent codes are selected by sampling randomly from
the normal distribution, with a ``truncation trick'' (Karras, Laine, and
Aila 2019, 4403) with a psi value 0.5.

\hypertarget{results}{%
\subsection{Results}\label{results}}

\hypertarget{finding-semantic-vectors}{%
\subsubsection{Finding semantic
vectors}\label{finding-semantic-vectors}}

Table 1 shows the accuracy of predictions made with a validation set
(balanced, including 20\% of the training set). The results show that
the features produced through projection have relatively high predictive
power relative to the IMD data; i.e., they contain information that
makes distinguishing between deprived and privileged areas possible. The
E4E and ReStyle models produce embeddings with more prediction power
than the SG2-ADA method. The health dimension was most strongly
connected with embeddings obtained with the SG2-ADA optimisation method.
The E4E embeddings are used in the later analysis in this paper since
they have the highest F1 score on most dimensions.

\begin{longtable}[]{@{}
  >{\raggedright\arraybackslash}p{(\columnwidth - 8\tabcolsep) * \real{0.2985}}
  >{\raggedright\arraybackslash}p{(\columnwidth - 8\tabcolsep) * \real{0.2687}}
  >{\raggedright\arraybackslash}p{(\columnwidth - 8\tabcolsep) * \real{0.1642}}
  >{\raggedright\arraybackslash}p{(\columnwidth - 8\tabcolsep) * \real{0.1194}}
  >{\raggedright\arraybackslash}p{(\columnwidth - 8\tabcolsep) * \real{0.1493}}@{}}
\caption{Precision, recall and F1 scores for three distinct dimensions
and inversion methods}\tabularnewline
\toprule\noalign{}
\begin{minipage}[b]{\linewidth}\raggedright
Dimension
\end{minipage} & \begin{minipage}[b]{\linewidth}\raggedright
Inversion method
\end{minipage} & \begin{minipage}[b]{\linewidth}\raggedright
Precision
\end{minipage} & \begin{minipage}[b]{\linewidth}\raggedright
Recall
\end{minipage} & \begin{minipage}[b]{\linewidth}\raggedright
F1 score
\end{minipage} \\
\midrule\noalign{}
\endfirsthead
\toprule\noalign{}
\begin{minipage}[b]{\linewidth}\raggedright
Dimension
\end{minipage} & \begin{minipage}[b]{\linewidth}\raggedright
Inversion method
\end{minipage} & \begin{minipage}[b]{\linewidth}\raggedright
Precision
\end{minipage} & \begin{minipage}[b]{\linewidth}\raggedright
Recall
\end{minipage} & \begin{minipage}[b]{\linewidth}\raggedright
F1 score
\end{minipage} \\
\midrule\noalign{}
\endhead
\bottomrule\noalign{}
\endlastfoot
Income & E4E & 0.794 & 0.721 & 0.756 \\
Education & E4E & 0.769 & 0.781 & 0.775 \\
Health & E4E & 0.899 & 0.754 & 0.820 \\
Income & ReStyle & 0.788 & 0.772 & 0.780 \\
Education & ReStyle & 0.773 & 0.763 & 0.768 \\
Health & ReStyle & 0.836 & 0.788 & 0.811 \\
Income & SG2-ADA & 0.716 & 0.738 & 0.727 \\
Education & SG2-ADA & 0.735 & 0.700 & 0.717 \\
Health & SG2-ADA & 0.747 & 0.738 & 0.742 \\
\end{longtable}

\hypertarget{visualising-latent-walks}{%
\subsubsection{Visualising latent
walks}\label{visualising-latent-walks}}

One option for systematically comparing changes in synthetic images is
to take a particular street view and see how it would change when
displaying features associated with low or~high health. Furthermore,
these changes could be compared with analogous manipulations related to
education and income. Image 1 shows a matrix created accordingly.
Notably, the initial image for this matrix is not a Street View image;
it is based on a randomly selected position in the model's latent space.

\begin{figure}

{\centering \includegraphics{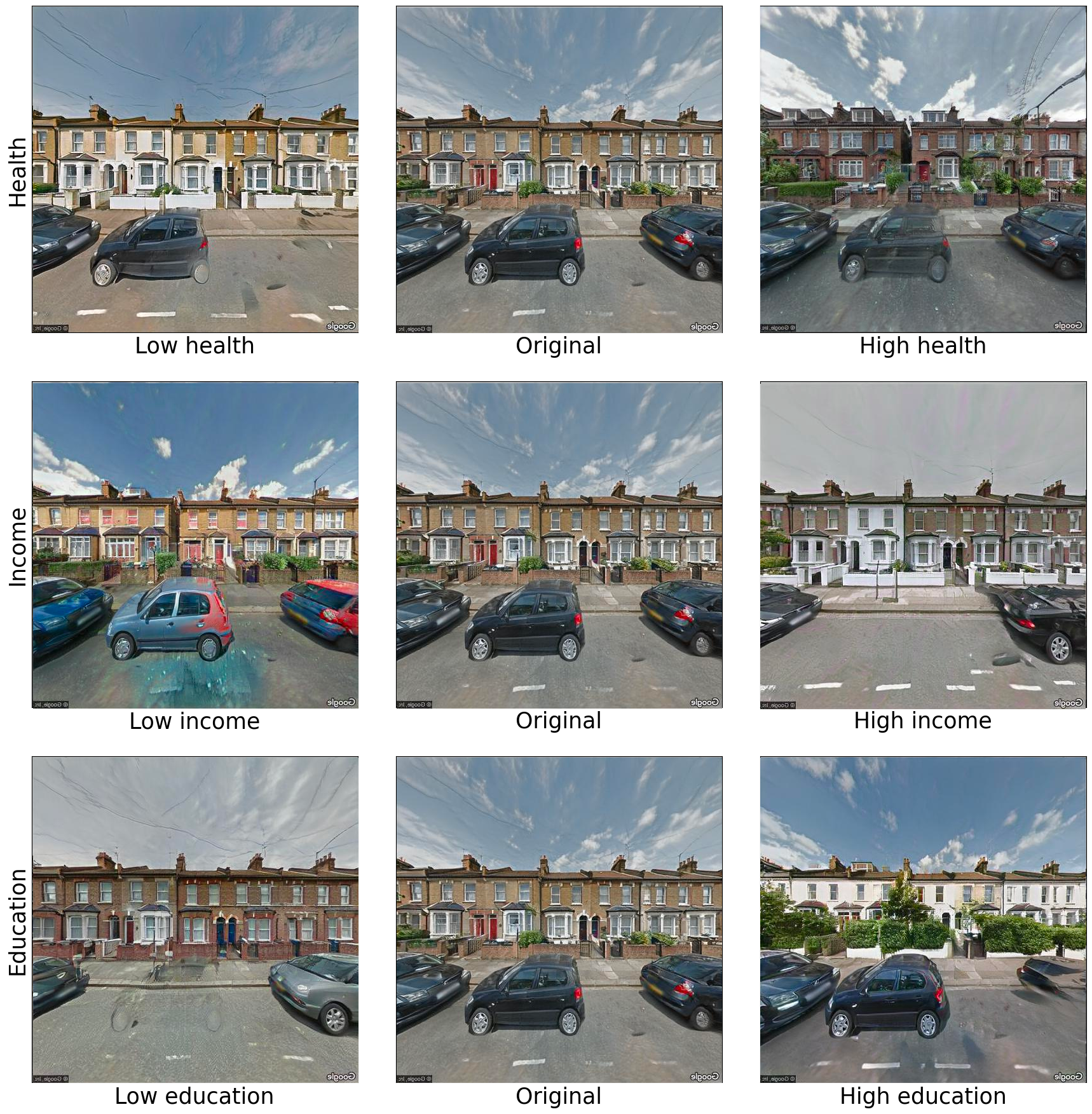}

}

\caption{Image 1: A matrix displaying a single synthetic image,
manipulated on three dimensions (health, income, and education)}

\end{figure}

Image 2 reflects an alternative way to organise a matrix, the first one
mentioned above. Instead of being arranged to present a single image
manipulated along several dimensions, it shows how several images change
along a single dimension (in this case, health). The manipulations
performed by the GAN are image-specific, so it is interesting to examine
whether particular dimensions exhibit effects that differ between, for
instance, two-floor semi-detached housing and three-floor variants of
what is known in the UK as terraced housing.

\begin{figure}

{\centering \includegraphics{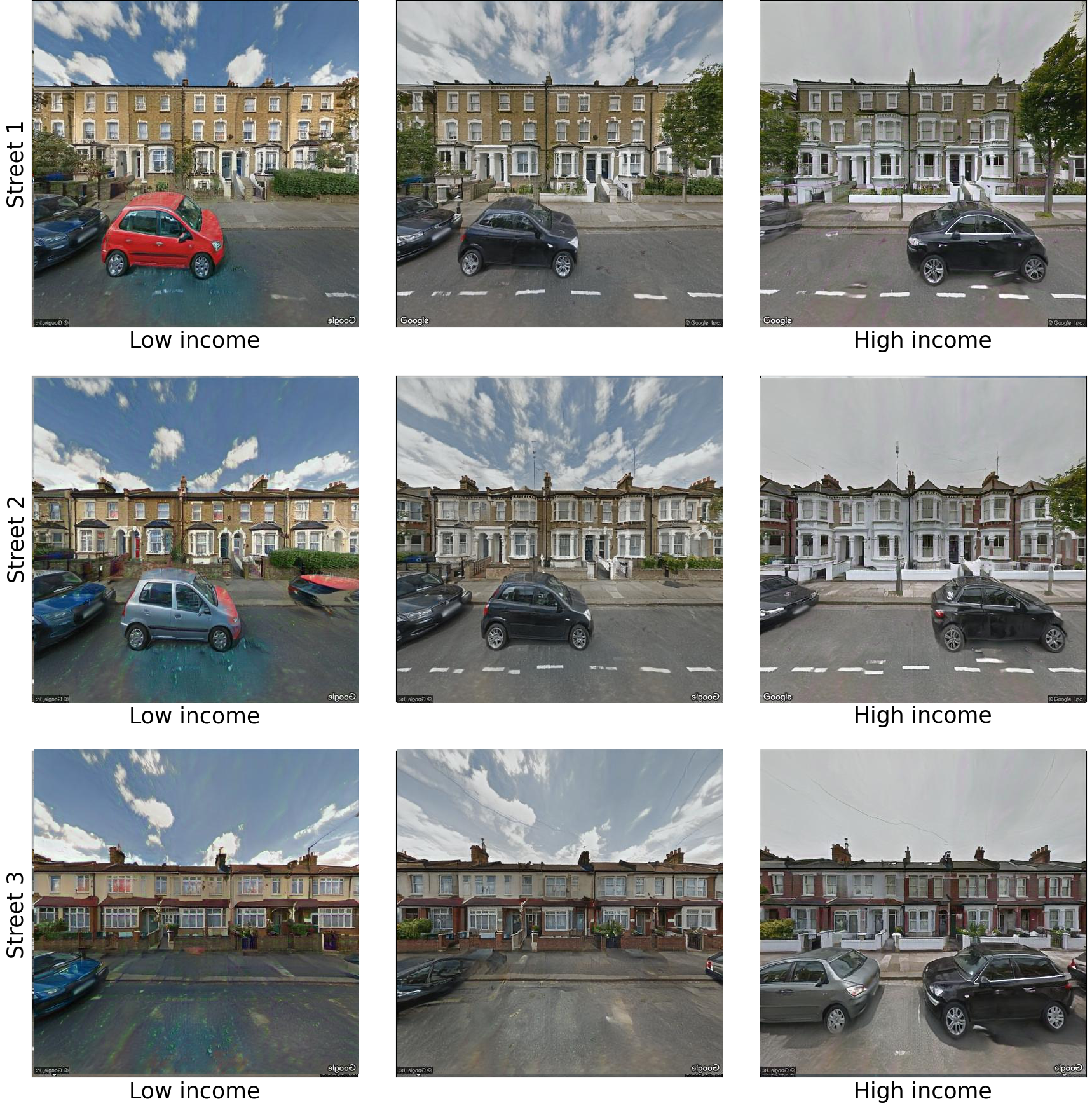}

}

\caption{Image 2. A matrix showing three distinct synthetic images
manipulated relative to the single dimension of health}

\end{figure}

Still, images reveal only one window on the results. Accordingly, the
demonstration in this paper is supplemented by an interactive website
whose interface lets the user direct the image synthesis via sliders.
The site, titled *``This Inequality Does Not Exist'', allows for
observing continuous image changes and combining adjustments on several
dimensions. It is accessible via
\url{http://knuutila.net/thisinequality}.

(Seta, Pohjonen, and Knuutila 2023) describe ``latent walks'' as a
systematic method of analysing visual features in synthetic images
created by models such as GANs.` Their paper also delves into what one
can discern about the visual features associated with deprivation and
privilege in London with the technique described in this paper. This
reading shows that synthetic images associated with education
deprivation contain 1930s and postwar building stock, while
higher-education areas tend to feature Victorian facades. With higher
education, roof pediments become more prevalent, and window frames grow
more intricately decorated. Thanks to hedges and bushes planted there,
front gardens are more secluded and private. A similar shift in visual
characteristics is visible when comparing lower-income neighbourhoods to
higher-income ones. Victorian bay windows give way to the structures of
Edwardian flat structures, and bare-brick buildings are often covered
with whitewashed stucco. Pavements transition from plain tarmac to
well-maintained paving stones. Conversely, turning one's gaze to
neighbourhoods with varying levels of health, the changes in the
street-view photographs relate less to greenery and more to how space is
utilised. Areas with better health indicators tend to have more
extensive front gardens and additional attic windows, pointing to loft
conversions for added living space. Likewise visible is a larger number
of trees lining the streets, with houses situated farther from vehicular
traffic. These visual shifts demonstrate well how specific architectural
features are associated with abstract phenomena such as education,
income, and health disparities between neighbourhoods.

\hypertarget{limitations}{%
\subsection{Limitations}\label{limitations}}

For the study, a subset of images from Google Street View was chosen in
line with specific criteria. With this subset reflected both in the
training set and in the images generated by the model, one type of
housing (terraced houses) features particularly prominently. At the same
time, there is a relative scarcity of other housing types, such as
housing estates: developments comprising detached single-family houses
and buildings with blocks of flats. This issue arises partly because
estates are underrepresented in the Street View data. The Google service
often does not cover the entrances to buildings of such types, so the
generative model predominantly characterises the visual aspects of
terraced housing. Since it may not accurately represent the full
spectrum of housing diversity in London, the model's ability to capture
and visualise associated socioeconomic variations, including deprivation
and privilege, may be limited by biased representation of the built
environment.

My approach relies on performing image inversion for all the images used
in training and validating the GAN model. Experimenting with various
inversion methods to uncover meaningful patterns in the data attests to
these methods' ability to generate embeddings that hold significant
power related to non-visual data derived from indices of multiple
deprivation domains. However, it is important to note that the results
of these methods vary, and each approach may introduce its own biases.
Additionally, the fidelity of the inverted images has limitations.
Certain crucial visual features in the original images cannot be
faithfully represented within the model's latent space.

This limitation is rooted in the design of the GAN architecture itself,
which was initially built for generating images from random vectors
rather than encoding arbitrary input images into the latent space. While
inversion methods continue to evolve and improve, this underlying
premise for the architecture constrains how detailed visual information
can be accurately captured in the latent space. Future studies could
focus more on comparing results between inversion methods. Also, a
similar approach for studying the visual qualities of social processes
through image generation could be followed with model architectures
other than GANs, which may be more conducive to this undertaking.

\hypertarget{conclusions}{%
\subsection{Conclusions}\label{conclusions}}

Generative models, be they GANs or more recently introduced models, such
as DALL-E and Stable Diffusion, have been lauded for their ability to
generate realistic images efficiently. My experiment with repurposing
generative models to visualise and develop knowledge demonstrated GAN
models' capacity to render tangible what Goetschalckx and colleagues
(2019) termed ``visual concepts'', referring to the visual correlates of
otherwise abstract qualities, such as disparities in income and
education. By conditioning the generation of synthetic urban images on
socioeconomic data, my framework sheds light on the relationship between
these abstract variables and visual aspects of urban environments. Thus,
it contributes a new perspective on the visual aspects of urban areas'
inequalities.

The project demonstrated the benefits of GAN models for this type of
analysis: GANs can recognise patterns within the training dataset and
map distinct visual attributes to specific subspaces within the model's
latent space. In combination, these two capabilities permit us to
generate images that exhibit gradual transitions aligned with these
visual attributes. Furthermore, they facilitate the disentanglement of
visual features related to variables such as education and income, even
when correlated with the source data.

While this approach has several limitations, it demonstrates solid
potential for exploiting generative models in visualisation and data
exploration. As large-scale visual datasets become more commonplace in
urban studies and many other fields, novel methods that utilise
generative models may help make patterns in the datasets or connections
with external data more tangible.

\newpage{}

\hypertarget{appendix-visual-comparison-of-projection-results-from-different-models}{%
\subsection{Appendix: Visual comparison of projection results from
different
models}\label{appendix-visual-comparison-of-projection-results-from-different-models}}

\begin{figure}

{\centering \includegraphics{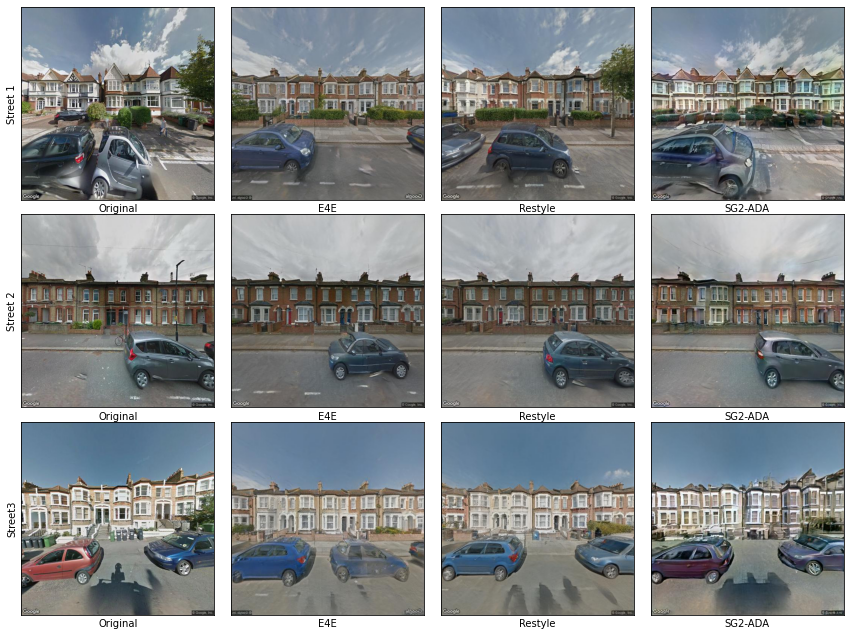}

}

\caption{Comparison of visual quality between projections of images made
by the ReStyle, Encoder4Editing, and StyleGAN2-ADA optimisation models}

\end{figure}

\hypertarget{references}{%
\subsection*{References}\label{references}}
\addcontentsline{toc}{subsection}{References}

\hypertarget{refs}{}
\begin{CSLReferences}{1}{0}
\leavevmode\vadjust pre{\hypertarget{ref-Alaluf2021}{}}%
Alaluf, Yuval, Or Patashnik, and Daniel Cohen-Or. 2021. {``ReStyle: A
Residual-Based StyleGAN Encoder via Iterative Refinement.''} In
\emph{Proceedings of the IEEE/CVF International Conference on Computer
Vision}, 6711--20.

\leavevmode\vadjust pre{\hypertarget{ref-Arietta2014}{}}%
Arietta, Sean M., Alexei A. Efros, Ravi Ramamoorthi, and Maneesh
Agrawala. 2014. {``City Forensics: Using Visual Elements to Predict
Non-Visual City Attributes.''} \emph{IEEE Transactions on Visualization
and Computer Graphics} 20 (12): 2624--33.
\url{https://doi.org/10.1109/TVCG.2014.2346446}.

\leavevmode\vadjust pre{\hypertarget{ref-Biljecki2021}{}}%
Biljecki, Filip, and Koichi Ito. 2021. {``Street View Imagery in Urban
Analytics and GIS: A Review.''} \emph{Landscape and Urban Planning} 215
(November): 104217.
\url{https://doi.org/10.1016/j.landurbplan.2021.104217}.

\leavevmode\vadjust pre{\hypertarget{ref-Chen2022}{}}%
Chen, Yizhou, Xu-Hua Yang, Zihan Wei, Ali Asghar Heidari, Nenggan Zheng,
Zhicheng Li, Huiling Chen, Haigen Hu, Qianwei Zhou, and Qiu Guan. 2022.
{``Generative Adversarial Networks in Medical Image Augmentation: A
Review.''} \emph{Computers in Biology and Medicine} 144 (May): 105382.
\url{https://doi.org/10.1016/j.compbiomed.2022.105382}.

\leavevmode\vadjust pre{\hypertarget{ref-LondonDatastore2022}{}}%
Datastore, London. 2022. {``LSOA Atlas.''}
https://data.london.gov.uk/dataset/lsoa-atlas.

\leavevmode\vadjust pre{\hypertarget{ref-Dorling2012}{}}%
Dorling, Daniel. 2012. \emph{The Visualisation of Spatial Social
Structure}. 2nd ed. Hoboken: John Wiley \& Sons.

\leavevmode\vadjust pre{\hypertarget{ref-Elasri2022}{}}%
Elasri, Mohamed, Omar Elharrouss, Somaya Al-Maadeed, and Hamid Tairi.
2022. {``Image Generation: A Review.''} \emph{Neural Processing Letters}
54 (5): 4609--46. \url{https://doi.org/10.1007/s11063-022-10777-x}.

\leavevmode\vadjust pre{\hypertarget{ref-Goetschalckx2019}{}}%
Goetschalckx, Lore, Alex Andonian, Aude Oliva, and Phillip Isola. 2019.
{``GANalyze: Toward Visual Definitions of Cognitive Image Properties.''}
arXiv. \url{https://doi.org/10.48550/arXiv.1906.10112}.

\leavevmode\vadjust pre{\hypertarget{ref-Gui2021}{}}%
Gui, Jie, Zhenan Sun, Yonggang Wen, Dacheng Tao, and Jieping Ye. 2021.
{``A Review on Generative Adversarial Networks: Algorithms, Theory, and
Applications.''} \emph{IEEE Transactions on Knowledge and Data
Engineering}, 1--1. \url{https://doi.org/10.1109/TKDE.2021.3130191}.

\leavevmode\vadjust pre{\hypertarget{ref-Johnson2020}{}}%
Johnson, Erik B., Alan Tidwell, and Sriram V. Villupuram. 2020.
{``Valuing Curb Appeal.''} \emph{The Journal of Real Estate Finance and
Economics} 60 (1): 111--33.
\url{https://doi.org/10.1007/s11146-019-09713-z}.

\leavevmode\vadjust pre{\hypertarget{ref-Kang2020}{}}%
Kang, Yuhao, Fan Zhang, Song Gao, Hui Lin, and Yu Liu. 2020. {``A Review
of Urban Physical Environment Sensing Using Street View Imagery in
Public Health Studies.''} \emph{Annals of GIS} 26 (3): 261--75.
\url{https://doi.org/10.1080/19475683.2020.1791954}.

\leavevmode\vadjust pre{\hypertarget{ref-Karras2019}{}}%
Karras, Tero, Samuli Laine, and Timo Aila. 2019. {``A Style-Based
Generator Architecture for Generative Adversarial Networks.''} arXiv.
\url{https://doi.org/10.48550/arXiv.1812.04948}.

\leavevmode\vadjust pre{\hypertarget{ref-Karras2020}{}}%
Karras, Tero, Samuli Laine, Miika Aittala, Janne Hellsten, Jaakko
Lehtinen, and Timo Aila. 2020. {``Analyzing and Improving the Image
Quality of StyleGAN.''} arXiv.
\url{https://doi.org/10.48550/arXiv.1912.04958}.

\leavevmode\vadjust pre{\hypertarget{ref-Kim2018}{}}%
Kim, Been, Martin Wattenberg, Justin Gilmer, Carrie Cai, James Wexler,
Fernanda Viegas, and Rory Sayres. 2018. {``Interpretability Beyond
Feature Attribution: Quantitative Testing with Concept Activation
Vectors (TCAV).''} arXiv.
\url{https://doi.org/10.48550/arXiv.1711.11279}.

\leavevmode\vadjust pre{\hypertarget{ref-Langer2020}{}}%
Langer, Tomas, Natalia Fedorova, and Ron Hagensieker. 2020.
{``Formatting the Landscape: Spatial Conditional GAN for Varying
Population in Satellite Imagery.''} arXiv.
\url{https://doi.org/10.48550/arXiv.2101.05069}.

\leavevmode\vadjust pre{\hypertarget{ref-Lei2020}{}}%
Lei, Na, Dongsheng An, Yang Guo, Kehua Su, Shixia Liu, Zhongxuan Luo,
Shing-Tung Yau, and Xianfeng Gu. 2020. {``A Geometric Understanding of
Deep Learning.''} \emph{Engineering} 6 (3): 361--74.
\url{https://doi.org/10.1016/j.eng.2019.09.010}.

\leavevmode\vadjust pre{\hypertarget{ref-Liu2023}{}}%
Liu, Ming, Yuxiang Wei, Xiaohe Wu, Wangmeng Zuo, and Lei Zhang. 2023.
{``Survey on Leveraging Pre-Trained Generative Adversarial Networks for
Image Editing and Restoration.''} \emph{Science China Information
Sciences} 66 (5): 151101.
\url{https://doi.org/10.1007/s11432-022-3679-0}.

\leavevmode\vadjust pre{\hypertarget{ref-Nasar1998}{}}%
Nasar, Jack L. 1998. \emph{The Evaluative Image of the City}. Thousand
Oaks, CA: Sage Publications.

\leavevmode\vadjust pre{\hypertarget{ref-Osokin2017}{}}%
Osokin, Anton, Anatole Chessel, Rafael E. Carazo Salas, and Federico
Vaggi. 2017. {``GANs for Biological Image Synthesis.''} In
\emph{Proceedings of the IEEE International Conference on Computer
Vision}, 2233--42.

\leavevmode\vadjust pre{\hypertarget{ref-Quercia2014}{}}%
Quercia, Daniele, Neil Keith O'Hare, and Henriette Cramer. 2014.
{``Aesthetic Capital: What Makes London Look Beautiful, Quiet, and
Happy?''} In \emph{Proceedings of the 17th ACM Conference on Computer
Supported Cooperative Work \& Social Computing}, 945--55. Baltimore
Maryland USA: ACM. \url{https://doi.org/10.1145/2531602.2531613}.

\leavevmode\vadjust pre{\hypertarget{ref-Salesses2013}{}}%
Salesses, Philip, Katja Schechtner, and César A. Hidalgo. 2013. {``The
Collaborative Image of the City: Mapping the Inequality of Urban
Perception.''} \emph{PLOS ONE} 8 (7): e68400.
\url{https://doi.org/10.1371/journal.pone.0068400}.

\leavevmode\vadjust pre{\hypertarget{ref-Schmidt2019}{}}%
Schmidt, Victor, Alexandra Luccioni, S. Karthik Mukkavilli, Narmada
Balasooriya, Kris Sankaran, Jennifer Chayes, and Yoshua Bengio. 2019.
{``Visualizing the Consequences of Climate Change Using Cycle-Consistent
Adversarial Networks.''} arXiv.
\url{https://doi.org/10.48550/arXiv.1905.03709}.

\leavevmode\vadjust pre{\hypertarget{ref-Seta2023a}{}}%
Seta, Gabriele de, Matti Pohjonen, and Aleksi Knuutila. 2023.
{``Synthetic Ethnography: Field Devices for the Qualitative Study of
Generative Models.''} SocArXiv.
\url{https://doi.org/10.31235/osf.io/zvew4}.

\leavevmode\vadjust pre{\hypertarget{ref-Shen2020}{}}%
Shen, Yujun, Ceyuan Yang, Xiaoou Tang, and Bolei Zhou. 2020.
{``InterFaceGAN: Interpreting the Disentangled Face Representation
Learned by GANs.''} arXiv. \url{https://arxiv.org/abs/2005.09635}.

\leavevmode\vadjust pre{\hypertarget{ref-UKNationalStatistics2019}{}}%
Statistics, UK National. 2019. {``Index of Multiple Deprivation
(IMD).''} \url{https://doi.org/10.20390/enginddepriv2015}.

\leavevmode\vadjust pre{\hypertarget{ref-Tonkiss2015}{}}%
Tonkiss, Fran. 2015. \emph{Space, the City and Social Theory: Social
Relations and Urban Forms}. Reprint. Cambridge: Polity.

\leavevmode\vadjust pre{\hypertarget{ref-Tov2021}{}}%
Tov, Omer, Yuval Alaluf, Yotam Nitzan, Or Patashnik, and Daniel
Cohen-Or. 2021. {``Designing an Encoder for StyleGAN Image
Manipulation.''} arXiv. \url{https://doi.org/10.48550/arXiv.2102.02766}.

\leavevmode\vadjust pre{\hypertarget{ref-Tufte1998}{}}%
Tufte, Edward R. 1998. \emph{Visual Explanations: Images and Quantities,
Evidence and Narrative}. 3rd. print., with revisions. Cheshire (Conn.):
Graphics Press.

\leavevmode\vadjust pre{\hypertarget{ref-Xia2022}{}}%
Xia, Weihao, Yulun Zhang, Yujiu Yang, Jing-Hao Xue, Bolei Zhou, and
Ming-Hsuan Yang. 2022. {``GAN Inversion: A Survey.''} arXiv.
\url{https://arxiv.org/abs/2101.05278}.

\end{CSLReferences}

\end{document}